# Electricity Consumption Forecasting: An Approach Using Cooperative Ensemble Learning with SHapley Additive exPlanations

**Eduardo Luiz Alba *, Gilson Adamczuk Oliveira, Matheus Henrique Dal Molin Ribeiro and Érick Oliveira Rodrigues**

Industrial & Systems Engineering Graduate Program (PPGEPS), Federal University of Technology-Parana (UTFPR), Via do Conhecimento, KM 01—Fraron, Pato Branco 85503-390, PR, Brazil; gilson@utfpr.edu.br (G.A.O.); mribeiro@utfpr.edu.br (M.H.D.M.R.); erickr@id.uff.br (É.O.R.)

* Correspondence: eduardoalba@alunos.utfpr.edu.br

**Abstract:** Electricity expense management presents significant challenges, as this resource is susceptible to various influencing factors. In universities, the demand for this resource is rapidly growing with institutional expansion and has a significant environmental impact. In this study, the machine learning models long short-term memory (LSTM), random forest (RF), support vector regression (SVR), and extreme gradient boosting (XGBoost) were trained with historical consumption data from the Federal Institute of Paraná (IFPR) over the last seven years and climatic variables to forecast electricity consumption 12 months ahead. Datasets from two campuses were adopted. To improve model performance, feature selection was performed using Shapley additive explanations (SHAP), and hyperparameter optimization was carried out using genetic algorithm (GA) and particle swarm optimization (PSO). The results indicate that the proposed cooperative ensemble learning approach named Weaker Separator Booster (WSB) exhibited the best performance for datasets. Specifically, it achieved an sMAPE of 13.90% and MAE of 1990.87 kWh for the IFPR–Palmas Campus and an sMAPE of 18.72% and MAE of 465.02 kWh for the Coronel Vivida Campus. The SHAP analysis revealed distinct feature importance patterns across the two IFPR campuses. A commonality that emerged was the strong influence of lagged time-series values and a minimal influence of climatic variables.

## 1. Introduction

The electricity consumption of educational institutions has a significant impact on the environment [1,2]. The demand for this resource in these institutions has been rapidly increasing due to rising student admissions, the introduction of new courses, and the growing number of research centers, which implies the need for more efficient and sustainable energy use [3]. Consequently, to maintain a competitive advantage, it is imperative to prioritize continuous enhancements in operational efficiency [4]. Effective energy consumption management further contributes to enhancing the institution's overall sustainability and competitiveness.

However, efficiently managing electricity consumption is a complex undertaking due to its vulnerability to a range of cost-impacting factors, including equipment malfunctions and power grid instabilities. Under these circumstances, accurately forecasting electricity patterns is crucial for optimizing resource allocation and operational planning [5].

Time series represent ordered sequences of observations collected over time [6]. Forecasting these series entails predicting future values based on historical patterns, traditionally employing statistical or machine learning (ML) methodologies [7]. Also, due to the number of features associated with real-time series, it is necessary to use improved forecasting models to achieve reliable forecasting accuracy. In this direction, using ensemble learning methods, especially cooperative ensemble learning models, is attractive. These approaches can combine feature engineering techniques, time-series decomposition, and optimization strategies for hyperparameter fine-tuning [8–10].

This study presents a novel cooperative ensemble learning method, the Weaker Separator Booster (WSB), comprising four machine learning models: random forest (RF), support vector regression (SVR), long short-term memory (LSTM), echo state networks (ESN), convolutional neural networks (CNN), and extreme gradient boosting (XGBoost). Hyperparameter optimization was conducted using genetic algorithms (GA) and particle swarm optimization (PSO). The proposed approach was applied to forecast electricity consumption 12 months ahead at an educational institution. Shapley additive explanations (SHAP) were employed for feature selection and to evaluate the influence of exogenous variables.

This paper contributes to the existing literature in several ways:

1. The first contribution lies in using several external variables in the forecasting system. These variables are related to calendar events, the pandemic, and climatic conditions. In this study, the SHAP analysis is adopted to perform feature selection and feature importance. In other words, SHAP analysis is conducted in the initial stage to select the most suitable set of inputs for the forecasting models. Next, the importance of each set of variables in achieving accurate electricity consumption forecasts is accessed through the SHAP values.
2. Second, different artificial intelligence models, including neural networks (LSTM, CNN, and ESN), ensemble methods (RF and XGBoost), support vector machines (SVR), and classical statistical models (exponential smoothing, Holt-Winters, and ARIMA), are employed. Optimization strategies such as GA and PSO are used to fine-tune the model hyperparameters and find the most appropriate set of hyperparameters for each adopted model. This aims to provide insights to the reader regarding the most suitable model, in terms of forecasting error reduction, for electricity consumption forecasting in educational institutions.
3. Third, a new cooperative ensemble learning model is proposed to combine the strengths of each compared forecasting model. Specifically, we proposed an approach named Weaker Separator Booster (WSB) functions as a method for increasing the gap between the best predictor (GA–LSTM) and the weaker ones (which likely struggled to learn effectively and, as a result, tend to produce predictions closer to a straight horizontal line). We average these weaker predictors following the no-free-lunch theorem, generating a less biased and more generalized weaker result.

The remainder of this study is structured as follows: Section 2 presents a review of related work on electricity consumption forecasting. Section 3 provides a detailed description of the materials and methods employed in the study. Section 4 presents and discusses the results obtained from the study. Finally, Section 5 summarizes the overall conclusions of the study and outlines proposals for future work.

## 2. Related Work

Generally, research on electricity consumption forecasting predominantly employs two methodologies: statistical models and machine learning techniques [11]. Regardless of the approach, model performance is typically assessed using specific error measures to quantify prediction accuracy and facilitate comparative analysis.

Focusing on machine learning models, a comprehensive review of the literature reveals a diversity of approaches employed for electricity consumption forecasting. This review sought out the most recent work related to the topic, utilizing the Scopus and Web of Science databases. A comprehensive search for recent developments in this field was conducted using the keywords “Machine Learning”, “Electricity”, and “Forecasting”, which yielded a large dataset of over 2000 studies. The results were then filtered based on the abstracts, with a focus on studies related to electricity forecasting in similar contexts, such as industrial and household applications. These encompass tree-based models, artificial neural networks, and support vector machines, as presented in Table 1.

**Table 1.** Reported studies for electricity consumption forecasting with ML approaches.

| Reference | Models | Criterion | Steps Ahead | Percentual Errors | Application |
|---|---|---|---|---|---|
| Abdelhamid et al. [12] | DTO-LSTM; STM; RF; SVM; KNN; MLP; Seq2Seq | RMSE; RRMSE; MAE; BEM; NSE; WI; $R^2$; R | 1 min | DTO-LSTM: 6.23% | Smart Households |
| Aabadi et al. [13] | PSO-XGBoost; XGBoost | MAE; MAPE; RMSE | 3 months | Cluster 1: 23.33%; Cluster 2: 31.00%; Cluster 3: 11.62%; Cluster 4: 23.11%; Cluster 5: 8.46% | Educational Institution |
| Bouktif et al. [14] | LSTM-GA; LSTM-PSO; Multi-seq; LSTM; RF; SVR; ANN; Extra Trees Regressor | MAE; RMSE; CV(RMSE) | – | – | Metropolitan |
| Kazmi1 et al. [5] | ANN; RF | MAE; MAPE; RMSE | 1 min; 15 min; 60 min | RF 01 Min: 2.42%; RF 15 Min: 3.70%; RF 60 Min: 4.62%; ANN 01 Min: 2.60%; ANN 15 Min: 5.20%; ANN 60 Min: 5.80% | Educational Institution |
| Li et al. [15] | KNN; NB; SVR; LMSR | MAPE; R; RMSE | 1 step | Electr-KNN: 5.90%; Electr-NB: 8.20%; Electr-SVR: 2.90%; Electr-LMSR: 2.60%; Heat-KNN: 10.10%; Heat-NB: 11.30%; Heat-SVR: 9.30%; Heat-LMSR: 11.30% | Building |
| Izidio et al. [16] | EvoHyS; SARIMA-MetaFA-LSSVR; SARIMA-PSO-LSSVR; SARIMA; MLP; SVR; LR; C & R Tree; SVR + LR; Bagging with MLPs | R; RMSE; MAE; MAPE; MaxAE | – | SARIMA-MetaFA-LSSVR: 15.66%; SARIMA-PSO-LSSVR: 16.19%; EvoHyS: 12.74% | Building |
| Ayub et al. [17] | CNN-GRU-EWO; SVM-GWO; SVM; CNN; LR; ELM | MAE; MAPE; MSE; RMSE; Accuracy; Precision; Recall; F1-Score | 7 days; 30 days | CNN-GRU-EWA: 6.00%; SVM-GWO: 1.33% | Metropolitan Area |
| Feng et al. [18] | LR; RT; SVM | $R^2$ | 1 step | – | Building |

The integration of optimization algorithms with ML algorithms is highly promising in time-series forecasting. The most prevalent application of optimization algorithms found in related literature pertains to optimizing the hyperparameters of algorithms, which directly impact their accuracy and generalization capability. Abdelhamid et al. [12] employed an optimization algorithm specifically for this purpose in the LSTM artificial neural network, demonstrating improved model effectiveness in forecasting energy consumption in smart households. Aabadi et al. [13] applied this approach to the XGBoost algorithm, using PSO, and observed enhanced performance in electricity consumption forecasting on an academic campus. Bouktif et al. [14] also utilized PSO and GA to improve forecast accuracy in electricity prediction for a metropolitan area using LSTM, achieving superior performance compared to other models. Finally, Kazmi et al. [5] proposed a triply optimized approach for predicting electricity consumption in university campuses, based on specific parameter selection for RF and artificial neural network (ANN) models.

Another approach to integrating optimization algorithms into time-series forecasting associated with ML models is using feature selection, where inadequate selection can compromise model performance and increase problem complexity. Li et al. [15] employed GA to select optimal features for predicting electricity demand in mixed-use buildings using KNN, SVR, naive Bayes, and linear minimum squared regression (LMSR) models, achieving significant performance gains. Izidio et al. [16] proposed a hybrid system integrating statistical techniques and ML with feature selection via GA for forecasting energy consumption using smart meters in a residential building. The models employed include ARIMA, SVR, LSTM, multilayer perceptron (MLP), linear regression (LR), and classification and regression trees, among others, which resulted in the best forecasts.

Approaches combining feature analysis/selection with hyperparameter optimization are also found in the literature. Ayub et al. [17] proposed a multi-model approach for electricity load forecasting using feature selection through RF and XGBoost algorithms with recursive feature elimination, and hyperparameter optimization of various ML models through grey wolf optimization (GWO) and evolutionary whale optimization (EWO), achieving excellent results. Feng et al. [18] also employed feature selection with hyperparameter optimization using SHAP values for forecasting energy consumption in buildings, demonstrating a positive correlation between hyperparameter optimization and model accuracy in a scenario with a reduced number of features.

Although various optimization applications for electricity consumption forecasting have been explored, few studies address their application in educational institutions using cooperative ensemble learning models. These institutions have unique characteristics that set them apart, such as fixed vacation periods, alternating activity schedules, and diverse available spaces. Additionally, variations in the number of occupants throughout the day and week, in addition to the occurrence of sporadic events like lectures and conferences, contribute to a more complex and variable consumption pattern. Due to this gap in the literature, this area appears to be promising for future research that could contribute to the implementation of more effective and sustainable energy management strategies.

## 3. Materials and Methods

In this section, the dataset used in the experiments is described. Additionally, the methods employed to develop electricity consumption forecasting are presented.

### 3.1. Dataset

The data were obtained from the electricity consumption (in kilowatt-hour—kWh) invoices of the Federal Institute of Paraná (IFPR) for two campuses: Palmas (dataset 1) and Coronel Vivida (dataset 2), encompassing 79 months, spanning September 2017 to March 2024. These data are available at https://github.com/eduardoalba0/ifpr_electricity_ml (accessed on 30 August 2024) and summarized in Table 2. Figure 1 illustrates the IFPR campuses map.

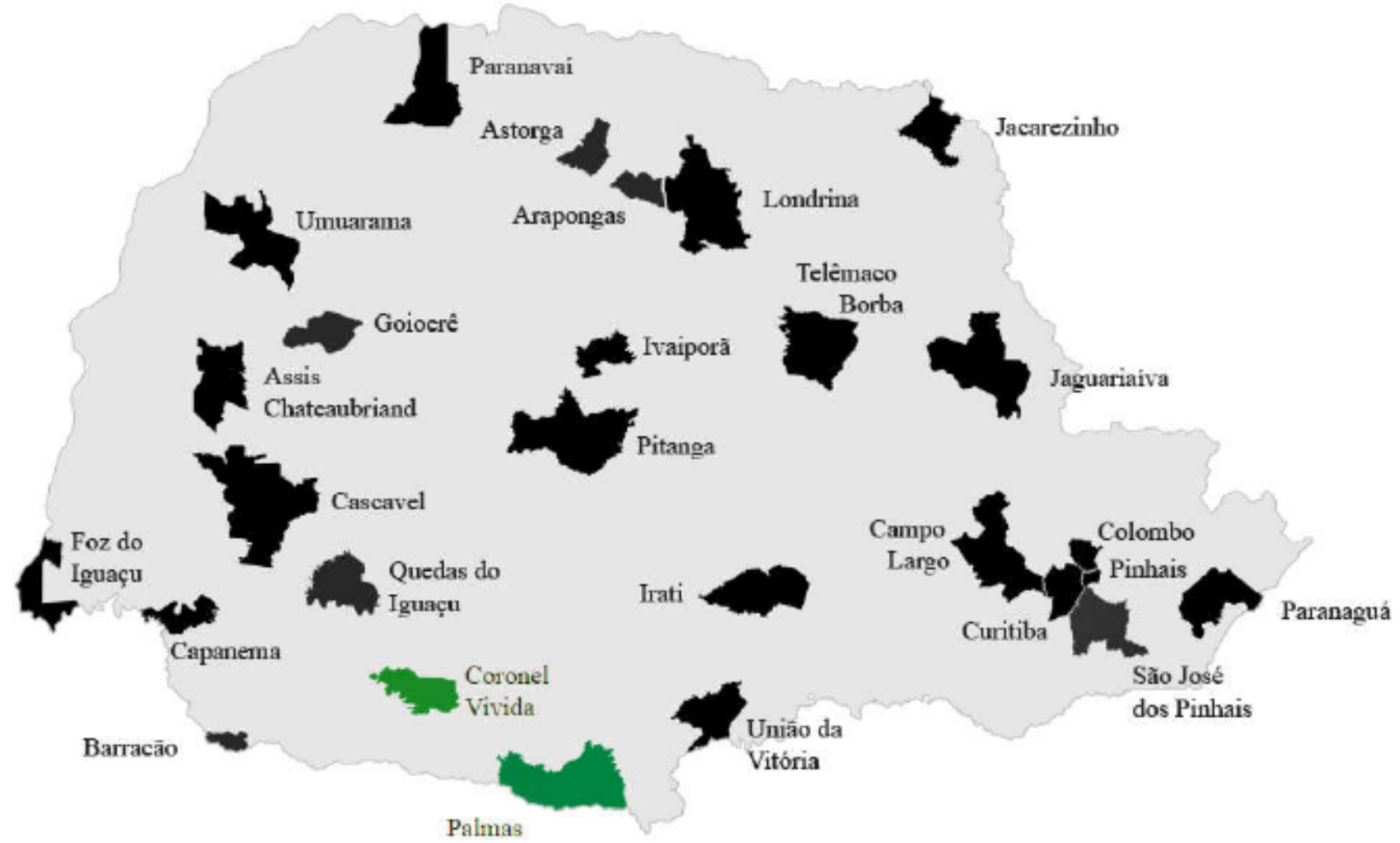

**Figure 1.** IFPR campus map: Green markers denote the campuses analyzed in this research.

**Table 2.** Statistical indicators (in kWh) for electricity consumption.

| Time Series | Observations | Minimum | Maximum | Median | Mean | Standard Deviation |
|---|---|---|---|---|---|---|
| Dataset 1 | 79 | 7252 | 25,339 | 15,342 | 15,995.58 | 4693.60 |
| Dataset 2 | 79 | 1087 | 4579 | 2182 | 2262.53 | 636.45 |

The IFPR is a Brazilian public federal educational institution dedicated to providing higher, basic, and professional education. Its academic offerings include both undergraduate and technical courses in diverse fields, available in the morning, afternoon, and evening. For this study, we selected two campuses with inversely proportional sizes but similar climatic conditions: the Palmas and Coronel Vivida campuses. The Palmas campus is the largest institution within IFPR, whereas Coronel Vivida is one of the smallest. Both are located in the southwestern region of Paraná State.

An analysis of the Palmas and Coronel campuses dataset, as depicted in Figure 2, reveals a highly irregular pattern in electricity consumption. Notably, several observations exhibit a decline in energy usage at the beginning of each year. This phenomenon is likely explained by the school holidays, which occur between December and January, when there is a significant decrease in the number of people in the institution.

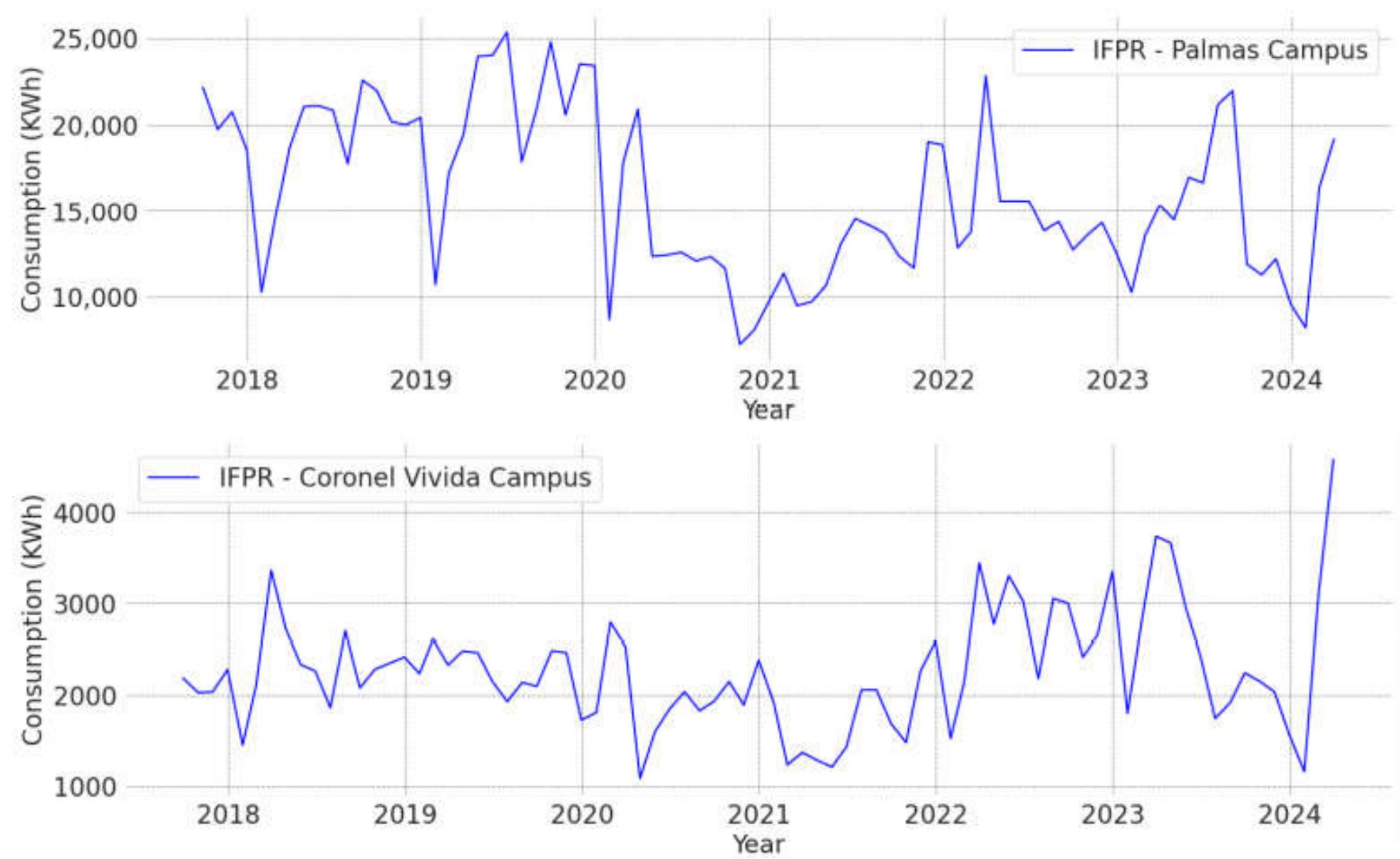

**Figure 2.** Monthly electricity consumption at dataset 1 (**top**) and dataset 2 (**bottom**).

The autocorrelation (ACF) and partial autocorrelation (PACF) functions were employed to analyze the correlations within the electricity consumption time series. As illustrated in Figure 3, these analyses revealed a degree of dependency between the data points and their lagged values for the Palmas Campus. This indicates that past electricity consumption values possess predictive power for current consumption levels.

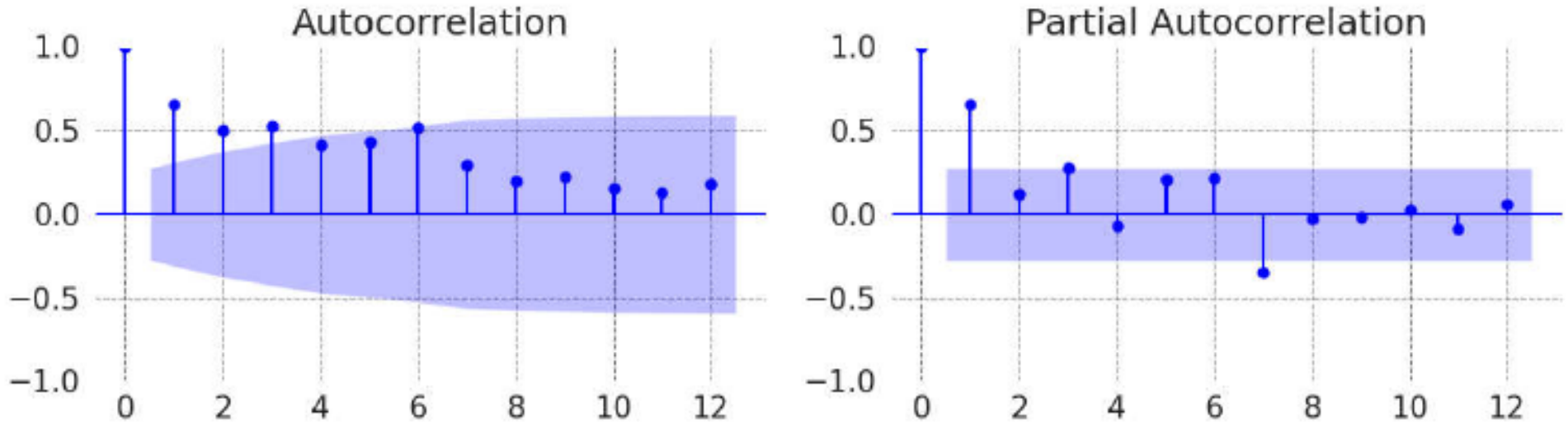

**Figure 3.** ACF and PACF dataset 1.

In contrast, in dataset 2, the effect of lagged values was less pronounced, with statistically significant correlations observed only at the first lags, as illustrated in Figure 4. This indicates that historical data have a weaker predictive power for current values compared to dataset 1.

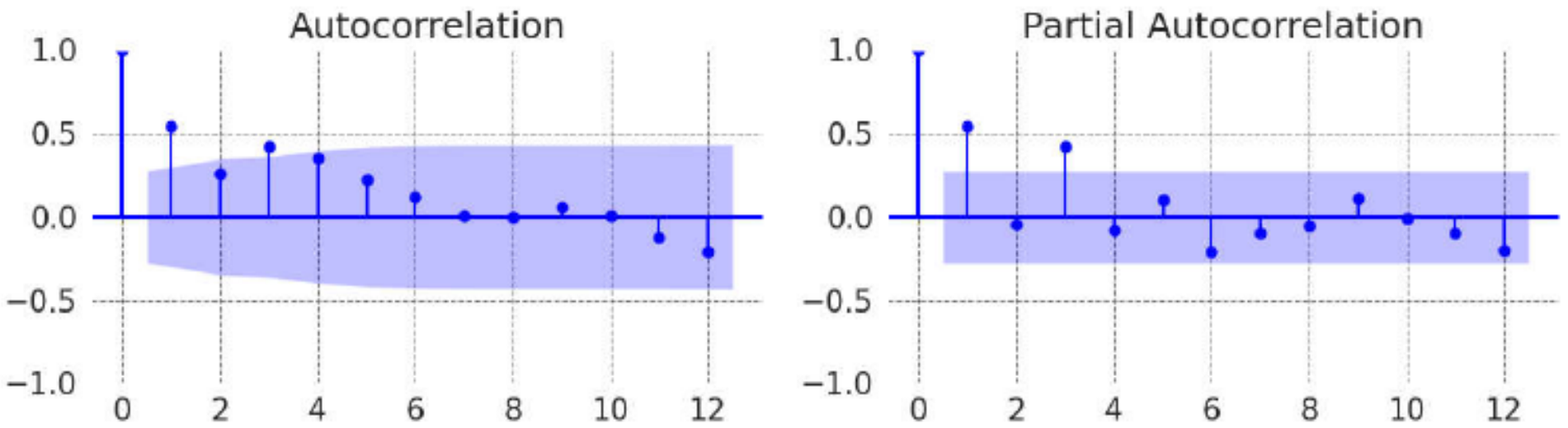

**Figure 4.** ACF and PACF of dataset 2.

The electricity consumption data at both campuses were found to be non-stationary and exhibit complex, irregular patterns, as indicated by hypothesis testing. This was especially evident in dataset 1, where neither trends nor seasonal patterns were detected, as summarized in Table 3.

**Table 3.** Summary of hypothesis test results.

| Time Series | Test | Statistics | Critical Value | Conclusion |
|---|---|---|---|---|
| Dataset 1 | KPSS | 0.1613 | 0.146 | There is no stationarity. |
| | Mann–Kendall | 0.0024 | 0.05 | There is no trend. |
| | Kruskal–Wallis | 0.4787 | 0.05 | There is no seasonality. |
| Dataset 2 | KPSS | 0.1902 | 0.146 | There is no stationarity. |
| | Mann–Kendall | 0.4067 | 0.05 | There is a trend. |
| | Kruskal–Wallis | 0.4777 | 0.05 | There is no seasonality. |

The KPSS test did not indicate stationarity in either campus, suggesting the absence of a unit root. Conversely, the Mann–Kendall test detected no significant trend in dataset 1 but indicated the presence of a trend in dataset 2, a trend which was not as clearly discernible in the visual representation of the series. The Kruskal–Wallis test confirmed the absence of seasonal patterns in both cases.

For the electricity consumption forecasting, exogenous variables were included in the dataset to enhance model training. The exogenous variables incorporated into the model are detailed in Table 4.

**Table 4.** Summary of exogenous variables incorporated into the models.

| Variable Set | Exogenous Variable | Frequency | Source |
|---|---|---|---|
| Climatic | Average Temperature | Monthly | IDR, SIMEPAR |
| | Max. Temperature—Absolute | Monthly | IDR, SIMEPAR |
| | Max. Temperature—Average | Monthly | IDR, SIMEPAR |
| | Max. Temperature—Average Absolute | Monthly | IDR, SIMEPAR |
| | Min. Temperature—Absolute | Monthly | IDR, SIMEPAR |
| | Min. Temperature—Average | Monthly | IDR, SIMEPAR |
| | Min. Temperature—Absolute Average | Monthly | IDR, SIMEPAR |
| | Rain | Monthly | IDR, SIMEPAR |
| | Relative Humidity | Monthly | IDR, SIMEPAR |
| | Solar Radiation | Monthly | IDR, SIMEPAR |
| | Wind Max. Speed | Monthly | IDR, SIMEPAR |
| | Wind Max. Speed—Absolute | Monthly | IDR, SIMEPAR |
| | Wind Max. Speed—Average | Monthly | IDR, SIMEPAR |
| | Wind Max. Speed— Absolute Average | Monthly | IDR, SIMEPAR |
| COVID | COVID-19 Pandemic Period Identifier | Monthly | IFPR |
| Date | Year | Monthly | – |
| | Month | Monthly | – |

Climate data, including temperature, precipitation, and wind speed, were obtained from the Paraná Institute of Rural Development (IDR-PR) and Paraná Meteorological System (SIMEPAR) for the period between 2018 and 2024. The inclusion of these climatic variables in the present study is supported by prior research, which has established their influence on electricity consumption, as indicated by Kazmi et al. [5], Li et al. [15], and Saxena et al. [19].

To mitigate the impact of the COVID-19 pandemic on model training, a pandemic period identifier was incorporated into the dataset. In-person activities at IFPR were suspended between March 2020 and April 2022, resulting in anomalous electricity consumption patterns. The inclusion of this identifier enabled the forecasting models to adapt to these atypical behaviors without compromising their accuracy during normal periods.

### 3.2. Proposed Methodology

The proposed methodology for predicting electricity consumption in educational institutions adheres to standard practices. The research process comprised three primary stages: data preparation and preprocessing, model training and testing, and performance evaluation, as depicted in Figure 5 and detailed in the following sections.

The proposed framework begins with a data preparation phase. These processes involved handling missing values and selecting relevant features through feature importance analysis based on the SHAP importance values. Less important features were removed to reduce model complexity and improve computational efficiency.

In the machine learning model setup phase, LSTM, RF, SVR, and XGBoost models were configured for forecasting. Hyperparameter optimization was conducted to identify adequate combinations of hyperparameters, aiming to enhance forecasting accuracy. The most adequate performing combination for each model was selected, resulting in four optimized models: GA–LSTM, GA–RF, GA–SVR, and GA–XGBoost.

The optimized models underwent a training process employing 6-fold cross-validation, followed by a testing phase encompassing a 12-month-ahead electricity consumption forecast. The performance of the proposed models was assessed using the MAE and sMAPE criteria. Subsequently, SHAP analysis was conducted to determine each incorporated variable's contribution and direction.

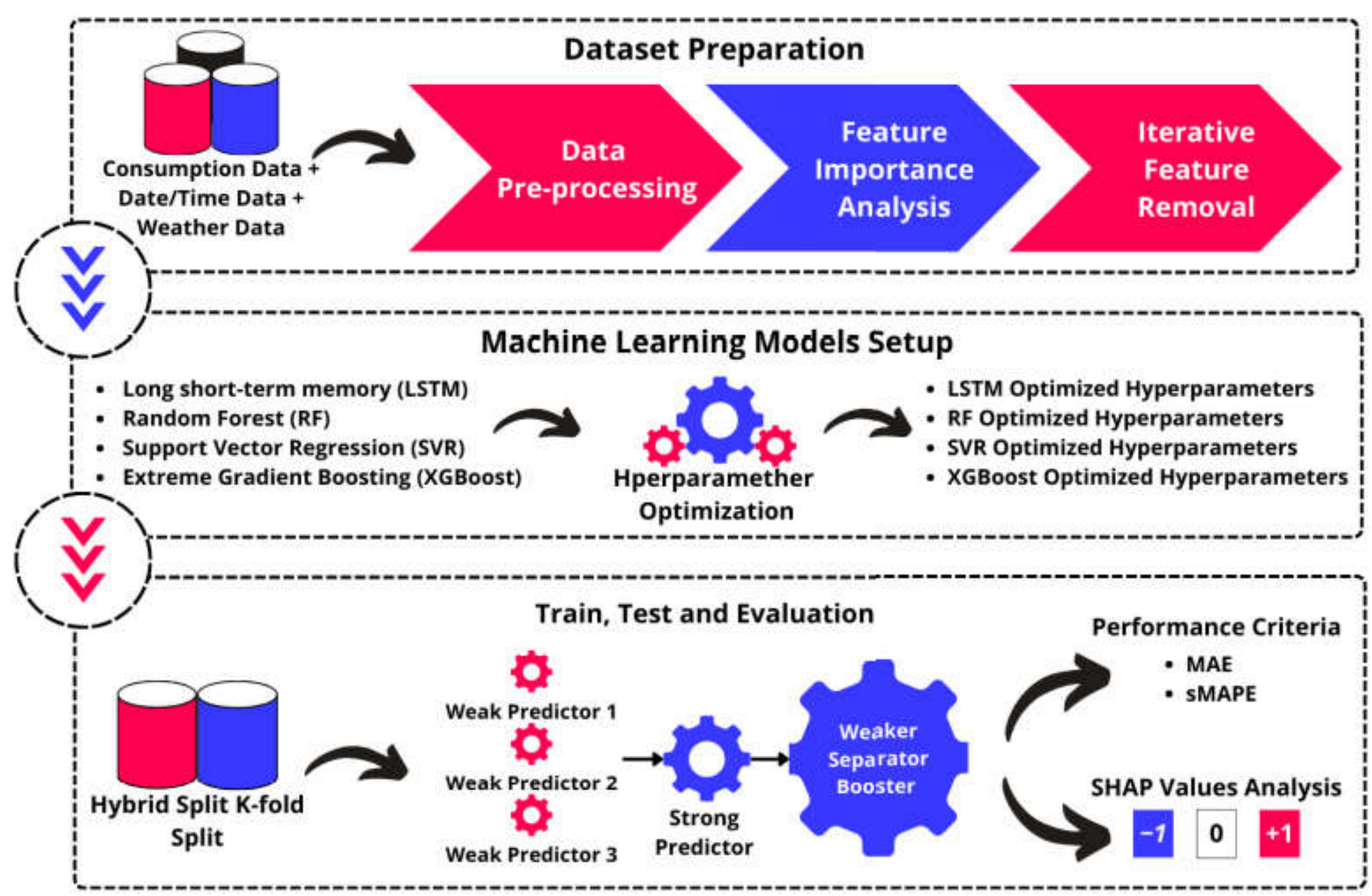

**Figure 5.** Proposed forecasting framework.

### 3.2.1. Dataset Preparation

The data preparation phase focused on refining and standardizing the dataset to optimize electricity consumption forecasting for the educational institution. This process encompassed data preprocessing, feature importance analysis, and the elimination of less significant features to enhance model efficiency.

The preprocessing stage involved data imputation and transformation. Missing values, exclusively identified in the January 2021 climate data due to meteorological station failures in Palmas-PR, as reported by IDR, were addressed by replacing them with the arithmetic mean of the available observations for the same month in other years. To capture temporal dependencies, twelve lags were incorporated for the electricity consumption variable. Categorical variables, Month and Year, were converted into dummy variables for model compatibility. Finally, observations with missing values across any column were eliminated from the dataset.

Following data preprocessing and integration, a dataset comprising 47 predictors was constructed. To enhance model parsimony and interpretability, feature importance analysis was conducted using the SHAP method on the Palmas Campus data due to its more complex and variable consumption patterns, which may represent the behavior of both institutions while reducing computational resource consumption in this step. Grounded in coalition game theory [20], SHAP determines the contribution of each feature to the model's prediction by equitably distributing the "payoff" among predictors [21].

The SHAP feature importance derived from the XGBoost and RF models is visually represented in Figure 6.

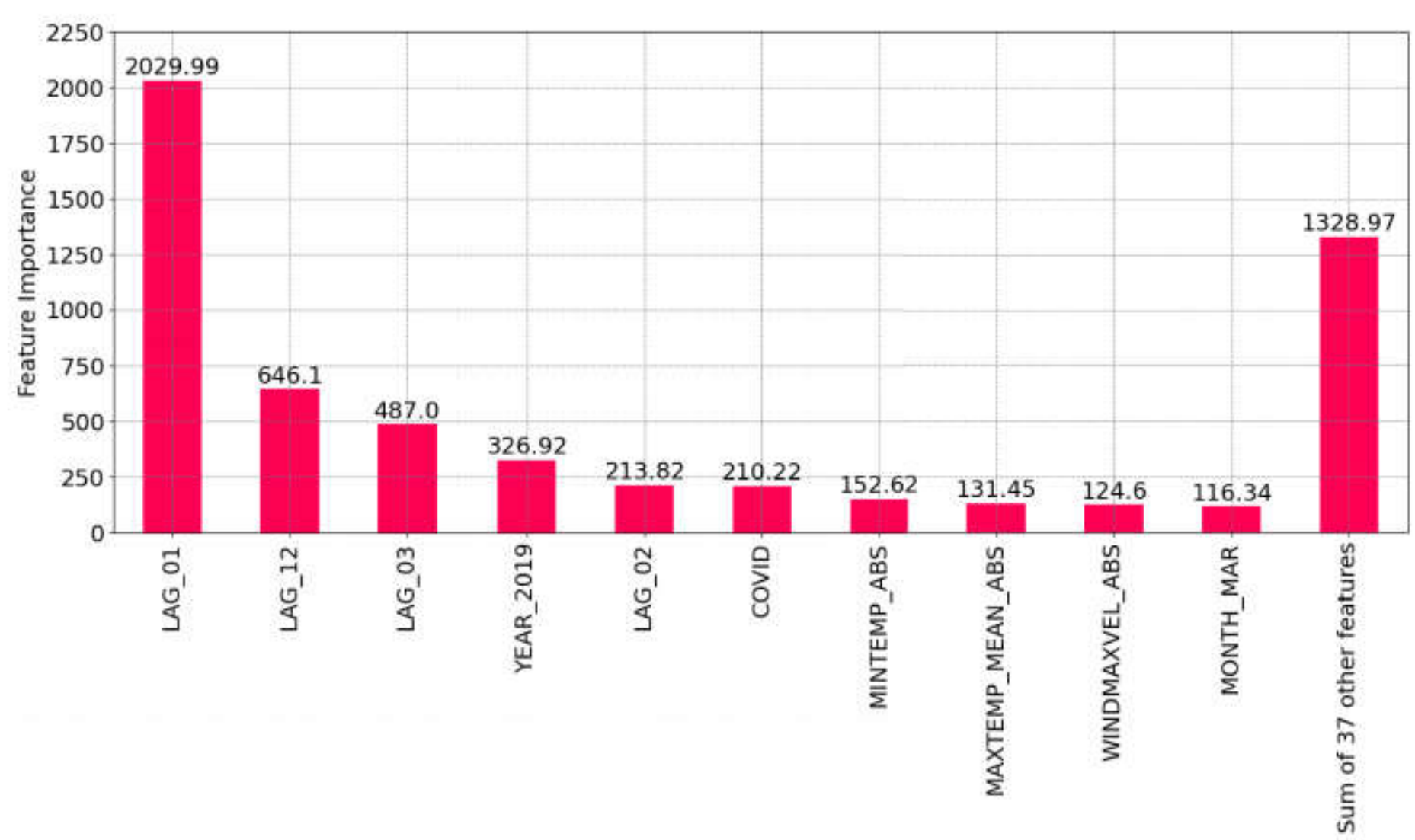

**Figure 6.** Average SHAP features importance between RF and XGBoost.

A sequential feature elimination process was implemented, ranked by increasing feature importance. At each iteration, the RF and XGBoost models were retrained, and their performance was assessed using the mean absolute error (MAE) with their GPU implementations. To prevent the removal of critical information and maintain model generalization, the feature elimination process was capped at 40 features. Figure 7 illustrates the MAE trajectory throughout the feature reduction process. The marked point corresponds to the minimum error between the average of the RF and XGBoost models.

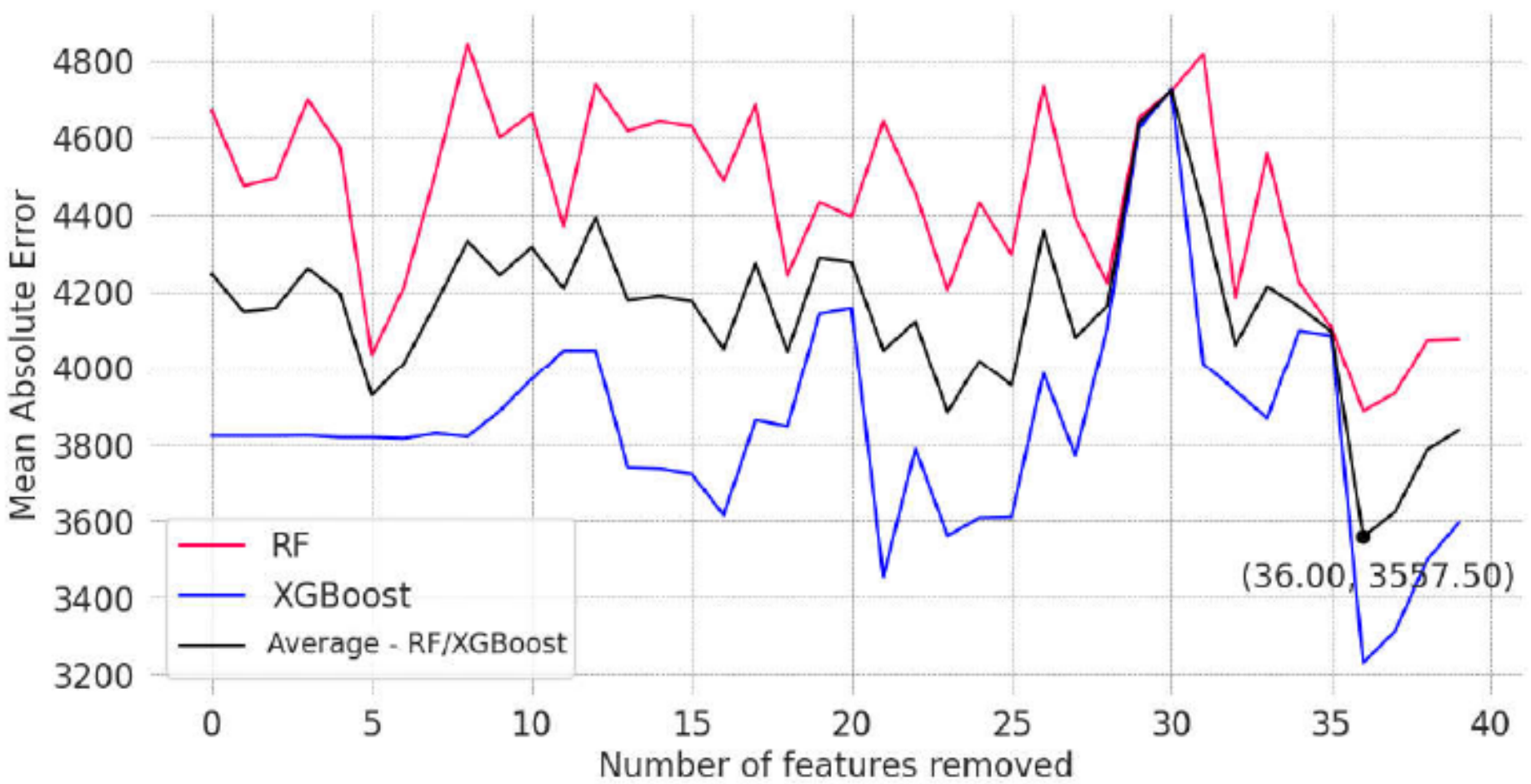

**Figure 7.** Comparison of MAE during the removal of less important features.

By analyzing the average performance of both the RF and XGBoost models, it was observed that the lowest MAE was achieved after removing the 36 least-important features from the dataset. The remaining 11 variables are listed in Table 5.

**Table 5.** The final set of features after SHAP analysis.

| Predictor Set | Predictor Name |
|---|---|
| Consumption Lags | Lag 1; Lag 2; Lag 3; Lag 6; Lag 12 |
| Climatic Data | Maximum Temperature—Average Absolute<br>Minimum Temperature—Absolute<br>Maximum Wind speed—Absolute |
| Date | Month—March<br>Year—2019 |
| COVID-19 | COVID-19 Pandemic Period Identifier |

#### 3.2.2. ML Models Setup

Given the extensive body of research on electricity consumption forecasting utilizing machine learning models, LSTM, RF, SVR, and XGBoost were selected for their widespread adoption and proven efficacy. The inclusion of these models, representing neural networks, decision trees, and support vector methods, respectively, enables a comprehensive comparative analysis of their predictive capabilities in this domain.

The LSTM model, a recurrent neural network employing a gradient-based learning algorithm [22], is widely adopted for time-series forecasting due to its capacity to preserve long-term dependencies [23]. The LSTM's core strength lies in its ability to efficiently transmit critical information across multiple time steps [24]. The hyperparameters employed in this LSTM model are detailed in Table 6.

**Table 6.** LSTM hyperparameters.

| Hyperparameter | Description | Type | Search Space | Selected Value |
|---|---|---|---|---|
| Units | Number of neurons in the LSTM layer | Integer | 1–300 | 115 |
| Epochs | Number of times the model iterates over the entire training dataset | Integer | 1–100 | 98 |
| Batch Size | Number of training samples used in each weight update | Integer | 1–300 | 117 |
| Activation | Function that determines a neuron output based on its weighted input | String | linear, mish, sigmoid, softmax, softplus, softsign, tanh | tanh |
| Bias | Constant term added to the weighted input before applying the activation function | Boolean | True, False | True |

The RF model is an ensemble method composed of multiple decision trees. Decision trees partition data into subsets based on simple decision rules. By aggregating the predictions of numerous decision trees, each trained on a random subset of the data, RF significantly improve predictive accuracy compared to a single decision tree [25]. The hyperparameter settings employed for the RF model in this study are outlined in Table 7.

**Table 7.** RF hyperparameters.

| Hyperparameter | Description | Type | Search Space | Selected Value |
|---|---|---|---|---|
| N_estimators | The number of trees in the forest | Integer | 10–300 | 13 |
| Max_depth | The maximum depth of each tree | Integer | 10–300 | 281 |
| Min_samples_split | The minimum number of samples required to split an internal node | Integer | 2–50 | 13 |
| Min_samples_leaf | The minimum number of samples required for a leaf node | Integer | 1–50 | 2 |

The SVR model is a supervised learning algorithm that identifies a function approximating the relationship between input variables and a continuous target variable while minimizing prediction error [26]. SVR employs support vectors, which are data points critical for defining the decision boundary, to construct a regression model. The model aims to find a function within a specified tolerance margin that best fits the observed data [27,28]. The hyperparameter configuration for the SVR model used in this study is presented in Table 8.

**Table 8.** SVR hyperparameters.

| Hyperparameter | Description | Type | Search Space | Selected Value |
|---|---|---|---|---|
| C | Regularization parameter controlling error penalty | Float | 0.00001–20,000.0 | 1465.0 |
| Epsilon | Tolerance for errors within a specified range | Float | 0.001–1.0 | 1.0 |
| Kernel | Determines the transformation used for finding separation hyperplanes | String | poly, rbf, sigmoid | sigmoid |

XGBoost is an ensemble method that employs gradient boosting with decision trees [29]. Similar to RF, XGBoost combines multiple decision trees to create a robust predictive model. However, unlike RF, XGBoost constructs trees sequentially, with each subsequent tree aiming to correct the errors of its predecessors. This approach, coupled with regularization techniques, enhances model generalization and mitigates overfitting [30,31]. The hyperparameter settings for the XGBoost model used in this study are outlined in Table 9.

**Table 9.** XGBoost hyperparameters.

| Hyperparameter | Description | Type | Search Space | Selected Value |
|---|---|---|---|---|
| N_estimators | The number of trees in the forest | Integer | 1–300 | 25 |
| Max_depth | The maximum depth of each tree | Integer | 1–300 | 82 |
| Booster | Type of booster used | String | gbtree, gblinear, dart | gbtree |
| Lambda | L2 regularization term on weights | Float | 0–100 | 0.151538 |
| Alpha | L1 regularization term on weights | Float | 0–100 | 97.963749 |

Convolutional neural network (CNN) is a feed-forward deep learning network comprising an input layer, hidden layers, pooling layers, fully connected layers, and output layers [32]. The hidden layers perform convolutions, which are operations applied to subregions of the input matrix, resulting in a convolutional matrix representing a feature value determined by the filter's coefficients and dimensions. These features are often more informative than the original features [33]. The convolutional layer is the most distinctive and crucial part of a CNN, as it extracts features from the input variables, integrates them, and maps them to output signals via the output layer [32]. Due to these characteristics, CNNs are widely used architectures in the deep learning community, particularly in computer vision tasks [34]. In our study, the hyperparameter settings for the CNN model are tabulated in Table 10.

**Table 10.** CNN hyperparameters.

| Hyperparameter | Description | Type | Search Space | Selected Value |
|---|---|---|---|---|
| Filters | The number of feature detectors (or filters) applied to each convolutional layer | Integer | 1–300 | 298 |
| Dense | The number of neurons in the dense layer | Integer | 1–500 | 276 |
| Epochs | Number of times the model iterates over the entire training dataset | Integer | 1–100 | 96 |
| Activation | Function that determines a neuron output based on its weighted input | String | linear, mish, sigmoid, softmax, softplus, softsign, tanh | softplus |

Echo state network (ESN) is a recurrent neural network with short-term memory capabilities, consisting of an input layer, a reservoir, and an output layer [35]. Their primary characteristic lies in a reservoir computing framework that connects all nodes from the input layer to the reservoir, which is subsequently connected to the output layer [36]. The internal layer (reservoir) comprises a large number of sparsely connected nodes. This sparse connectivity enables information to be propagated to other nodes, forming a short-term memory, making this model suitable for various applications such as time series and energy forecasting [37]. The hyperparameter settings for the ESN model used in this study are outlined in Table 11.

**Table 11.** ESN hyperparameters.

| Hyperparameter | Description | Type | Search Space | Selected Value |
|---|---|---|---|---|
| Reservoirs | Number of neurons in the reservoir layer | Integer | 2–1000 | 386 |
| Sparsity | Proportion of connections between reservoir neurons that are maintained | Float | 0.1–0.5 | 0.30 |
| Spectral Radius | Spectral radius of the recurrent weight matrix | Float | 0.1–0.9 | 0.54 |
| Noise | Noise added to each neuron (regularization) | Float | 0.0001–0.8 | 0.60 |

To optimize hyperparameter configurations for the selected machine learning models, both GA and PSO were employed. The objective function for both metaheuristic algorithms was to minimize the MAE of each model, using the Palmas Campus dataset. The hyperparameters were selected from the tuning options available in the utilized libraries and based on a review of related work.

GA is a metaheuristic optimization technique inspired by Darwinian evolution. It employs mathematical implementations of genetic operators, including crossover, mutation, and selection [38]. The GA begins with a randomly initialized population of individuals subjected to iterative modifications through the application of genetic operators. Crossover involves the combination of genetic material from two parent individuals to create offspring. Mutation introduces random alterations to the offspring, enabling search space exploration. The selection process determines the survival of individuals based on their fitness, or performance, in the given problem [39].

PSO is a metaheuristic algorithm inspired by the collective behavior of birds, flocks, and fish schools. A population of particles, each representing a potential solution, is iteratively updated. The velocity of each particle is adjusted based on its current position, personal best position, and the global best position within the swarm. This dynamic interplay between exploration (searching new regions) and exploitation (refining promising solutions) guides the swarm towards optimal regions of the search space [40–42].

A custom Python implementation of the GA, adhering to standard genetic operators, was developed for this study. To ensure parity with the PSO algorithm, which generates 30 new solutions per iteration, the GA was configured to produce 30 offspring per crossover generation. A specialized library was utilized for the PSO implementation. Detailed configurations for both optimizers are presented in Table 12.

**Table 12.** GA and PSO configuration details.

| Optimization Approach | Options | New Solutions per Iteration | Number of Iterations | Total Feature Evaluations |
|---|---|---|---|---|
| GA | Mutation Rate: 0.5 | 30 | 25 | 750 |
| PSO | Cognitive Param.: 0.5<br>Social Param.: 0.3<br>Swarm Inertia: 0.9 | 30 | 25 | 750 |

To expedite the evaluation of the objective function, GPU-accelerated implementations of LSTM, RF, SVR, and XGBoost were leveraged. To optimize computational efficiency, parallel processing threads were employed to concurrently assess the 30 individuals within each iteration of the optimization algorithms. For each individual (representing a specific hyperparameter configuration), a simple 5-fold cross-validation was conducted to estimate model performance, similar to the methodology adopted by Ribeiro et al. [43]. The final objective function value was computed by averaging the MAE obtained from each fold.

Three independent experimental runs were conducted, each employing a unique random seed. For both GA and PSO, the optimization process comprised 25 iterations or generations, with 30 individuals or particles per iteration. Consequently, a total of 750 objective function evaluations were performed for each algorithm and random seed. The average MAE for each optimization algorithm was recorded and compared across iterations, as shown in Figure 8.

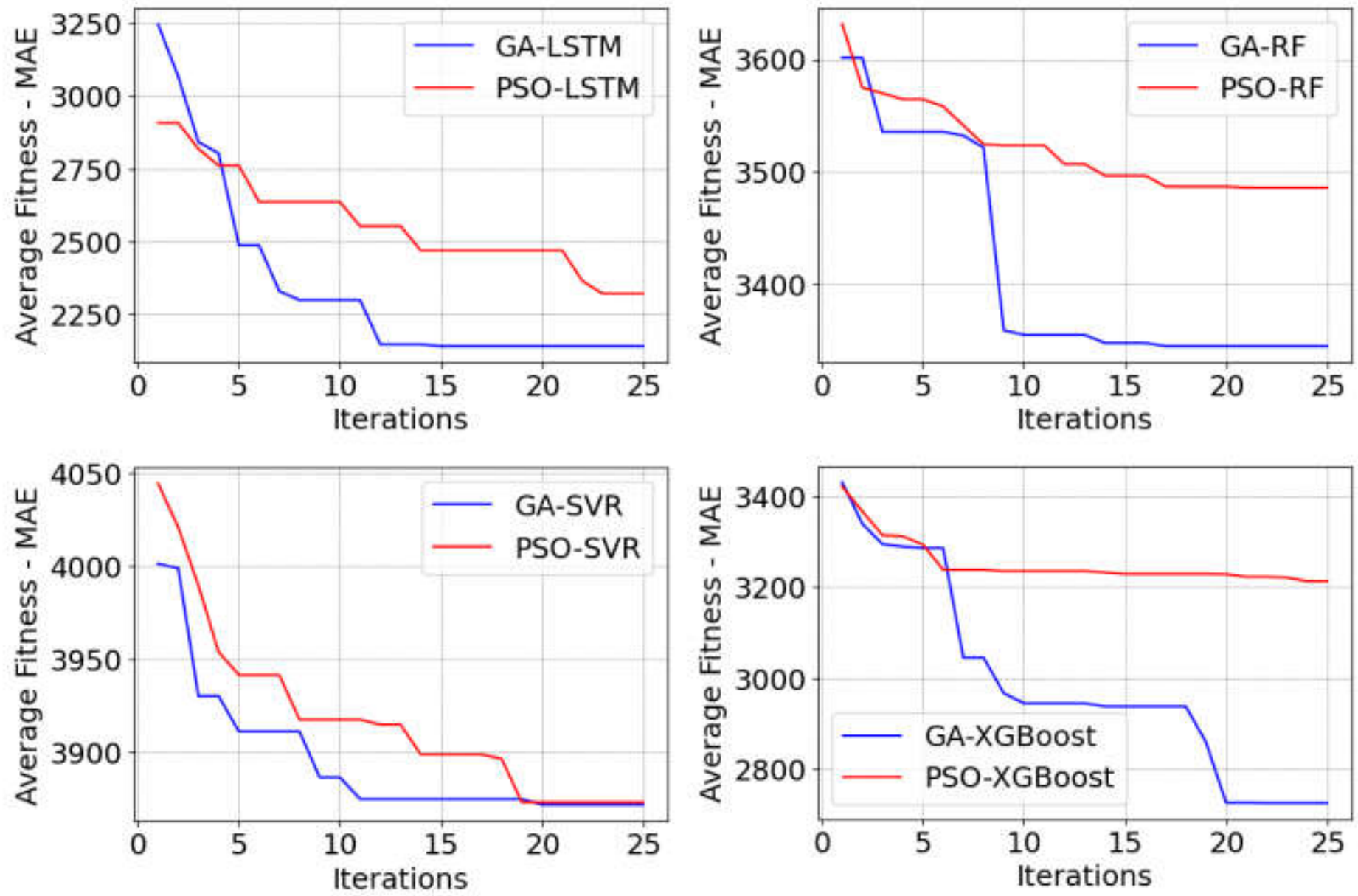

**Figure 8.** Comparison of fitness evolution between GA and PSO for LSTM, RF, SVR, and XGBoost hyperparameter tuning.

An analysis of the average fitness of the seeds for each optimization algorithm revealed that the GA found the best hyperparameter configurations for all models. The PSO experienced convergence difficulties for the RF and XGBoost models, becoming stuck in sub-optimal solutions. For the LSTM and SVR models, although the PSO converged more slowly, it approached the optimum found by the GA significantly, especially with the SVR model.

#### 3.2.3. Train, Test, and Evaluation

A hybrid data-splitting approach was employed for model evaluation. Initially, a temporal split was introduced, reserving the final 12 periods for the test set, aligning with the desired 12-month forecast horizon. The remaining data were subjected to 5-fold cross-validation for model training, as shown in Figure 9. This methodology enhanced the realism of performance assessment by simulating real-world forecasting conditions.

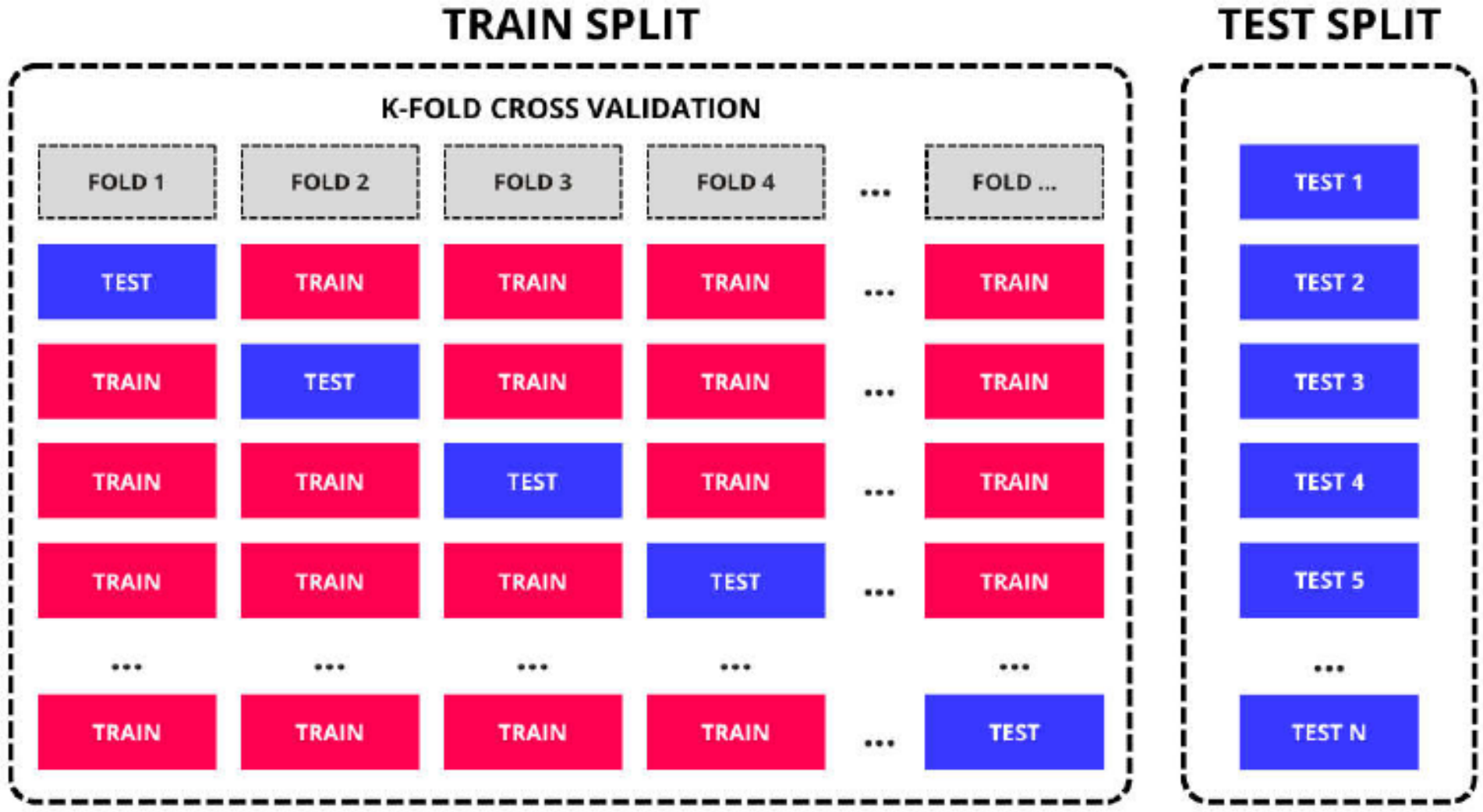

**Figure 9.** Hybrid split K-fold training and testing approach. The N refers to the forecast horizon.

During the training phase, each model was exposed to a comprehensive dataset encompassing both predictor variables and corresponding consumption outcomes. This allowed the models to learn the relationships between the variables and the target variable. Subsequently, in the testing phase, the models were evaluated on a distinct dataset that only contained predictor variables. The accuracy of the model's predictions, generated for this test set, was assessed by comparing them to the actual observed consumption values using appropriate error measures.

Experiments revealed that the GA–RF, GA–SVR, and GA–XGBoost models tend to generate simpler, linear models, which may indicate struggle to learn and to understand the patterns. Models tend to produce a straight line, as it reduces the error measure if placed in a reasonable position. However, the time series of electricity consumption presented in this study exhibited high variability and non-linear patterns, demanding more sophisticated models to capture their complexities. In this direction, we propose the WSB (Weaker Separator Booster) as a strategy to enhance the forecasts of the best model, the GA–LSTM.

The WSB can operate as two possible alternatives: (1) combining the LSTM ability to model more complex patterns with the stability of weaker models, or (2) improving the distance of the best model from the remaining of the weaker predictors, as this may disrupt the model and force it in the opposite direction of the inability to learn. In our approach, we focused on the second case due to the disposition of the predictions.

The WSB approach employs the average forecast of the GA–RF, GA–SVR, and GA–XGBoost models as a stable reference point of "not able to learn". This average, representing a more conservative prediction less susceptible to time series fluctuations, which conforms to the no-free-lunch theorem, is used to adjust the forecast generated by the GA–LSTM model, which was the best predictor in our experiments. An adaptive boosting mechanism, coupled with a weight parameter, determines, respectively, the value of the adjustment and its influence on the final GA–LSTM prediction, as detailed in Equations (1)–(3):

$$F_{WSB}(t) = F_s(t) + b(t) \tag{1}$$

$$b(t) = \left[F_s(t) - \frac{1}{n}\sum_{i=1}^{n} F_i(t)\right] \cdot w(t) \tag{2}$$

$$w(t) = \left|1 - \frac{|t_i - \frac{h}{2}|}{\frac{h}{2}}\right| \cdot w \tag{3}$$

where:

$F_{WSB}(t)$ is the enhanced prediction in time $t$;
$F_s(t)$ is the stronger model (GA–LSTM) prediction in time $t$;
$b(t)$ is the booster value in time $t$;
$n$ is the number of weaker models ($n$ = 3);
$F_i(t)$ is the weaker model (GA–SVR, GA–XGBoost, GA–RF) prediction in time $t$;
$t_i$ is the $i$th iteration;
$h$ is the prediction horizon ($h$ = 12);
$w(t)$ is the weight parameter in time $t$;
$w$ is the global weight parameter.

Given the recursive forecasting approach commonly employed in machine learning models, we observed a tendency for these models to accumulate errors from previous predictions. This accumulation often leads to larger errors as the forecast horizon increases. To mitigate this issue, we introduce an adaptive weighting scheme, detailed in Equation (3). This scheme dynamically adjusts the global weight at each iteration, starting with lower values and gradually increasing until an inflection point (at the midpoint of iterations), after which the weights decrease.

Different values were tested for the global weight variable, ranging from 0.0 to 1.0, with the optimal value found to be 0.0833 according to the random search approach. Higher values resulted in predictions that were either excessively larger or smaller than ideal, whereas smaller values had no significant effect on the model.

Subsequently, the model's performance was evaluated using several performance measures commonly employed in the literature. Traditionally, the MAPE criterion is usually employed in time-series error evaluation. However, it is biased towards lower forecasts and unsuitable for models expecting large errors. In this context, to overcome these drawbacks, the sMAPE can be adopted [44]. This study adopted the $MAE$ and $sMAPE$ as evaluation criteria, as defined in Equations (4) and (5):

$$MAE = \frac{1}{n}\sum_{i=1}^{n} |y_i - \hat{y}_i| \tag{4}$$

$$sMAPE = \frac{1}{n}\sum_{i=1}^{n} \frac{|y_i - \hat{y}_i|}{(|y_i| + |\hat{y}_i|)/2} \tag{5}$$

where:

$y_i$ are the observed values;
$\hat{y}_i$ is the forecast;
$n$ is the time-series length.

To gain insights into the feature importance and their impact on model predictions, a SHAP analysis was conducted. This analysis quantified the contribution of each variable to the model's output, revealing their relative importance and the direction of their impact (positive or negative). By understanding these relationships, the most critical variables and their interactions in shaping the final predictions were identified.

Experiments were conducted using Python Language v3.11.9 on an Acer Nitro 5 AN515-54 notebook equipped with an Intel Core i5-9300H CPU (2.40 GHz), 16 GB of RAM, and an NVIDIA GeForce GTX 1650 graphics card. The operating system was Windows 11 (×64) using the Windows Subsystem for Linux with Ubuntu 22.04. The Python libraries utilized in this study included: "pandas" and "numpy" for dataset preparation; "statsmodels" for time-series analysis and exponential smoothing; "pmdarima" for ARIMA forecasting; "pyESN" for echo state network model, "scikit-learn" for model training and testing; "dask" for creating multi-processing threads; "TensorFlow", "Keras.Conv1D", "Keras.LSTM", "XGBoost", and "cuML" for GPU-optimized models; "shap" for SHAP analysis; and "pyswarms" for PSO.

## 4. Results and Discussion

### *4.1. Model Evaluation*

Table 13 shows the results of 5-fold cross-validation training on four machine learning models for electric energy consumption forecasting: LSTM, RF, SVR, and XGBoost. The performance measures reported are MAE, median of MAE, and standard deviation of MAE.

**Table 13.** Statistics of 5-fold cross-validation training models.

| Time Series | Criteria | GA–LSTM | GA–RF | GA–SVR | GA–XGBoost |
|---|---|---|---|---|---|
| Dataset 1 | MAE (kWh) | 3930.20 | 1052.60 | 758.80 | 1879.20 |
| | Median of MAE (kWh) | 2691.00 | 634.00 | 941.00 | 2027.00 |
| | SD of MAE (kWh) | 2329.94 | 971.01 | 242.42 | 870.95 |
| Dataset 2 | MAE | 492.60 | 464.00 | 1458.80 | 441.20 |
| | Median of MAE | 451.00 | 640.00 | 1528.00 | 507.00 |
| | SD of MAE | 380.82 | 304.82 | 400.53 | 224.71 |

Based on the MAE and standard deviation, the GA–SVR model exhibited the lowest error during 5-fold cross-validation for the Palmas Campus time series, with an MAE of 758.80 and a standard deviation of 242.42. The GA–RF model followed, significantly behind, with an MAE of 758.80 and a standard deviation of 971.01. Considering the median, the GA–RF model presented the best value, with a median of 634.00, followed significantly by the GA–SVR model with a median of 941.00. Given the lower standard deviation, the GA–SVR model demonstrated more consistent performance scores in the training step for dataset 1.

In dataset 2, the GA–XGBoost model achieved the best scores in terms of mean and standard deviation during training, with an MAE of 441.20 and an SD of 224.71, followed by the GA–RF model with an MAE of 464.00 and an SD of 304.82. When considering the median, the GA–LSTM model presented the lowest error, with a value of 451.00, followed by the GA–XGBoost model. However, considering the lower standard deviation, the GA–XGBoost model exhibited the best training scores, presenting smaller fluctuations in values across each training fold for dataset 2.

In the testing phase, model performance was evaluated in a simulated real-world scenario, involving forecasting electricity consumption for 12 consecutive months at both the Palmas and Coronel Vivida campuses. To assess model performance, the MAE and sMAPE performance measures were employed, as shown in Tables 14 and 15.

Table 14. Performance measures for the compared models for dataset 1.

| **Proposed Models** | | | | | |
|---|---|---|---|---|---|
| **Criterion** | **WSB** | **GA–LSTM** | **GA-RF** | **GA–SVR** | **GA–XGBoost** |
| MAE (kWh) | 1990.87 | 2264.50 | 3211.33 | 3247.42 | 3760.75 |
| sMAPE (%) | 13.90 | 15.35 | 21.83 | 22.12 | 24.45 |
| Traditional Time Series Models | | | | | |
| **Criteria** | **Exponential Smoothing** | **Double Exponential Smoothing** | **Holt-Winters Additive** | **Holt-Winters Multiplicative** | **ARIMA** |
| MAE (kWh) | 3207.78 | 4190.56 | 5415.74 | 4534.49 | 3322.17 |
| sMAPE (%) | 21.82 | 30.10 | 43.31 | 33.84 | 22.65 |
| Other Artificial Neural Network Models | | | | | |
| **Criteria** | **GA–CNN** | **GA–ESN** | | | |
| MAE (kWh) | 5083.75 | 3951.25 | | | |
| sMAPE (%) | 39.03 | 26.30 | | | |

Table 15. Performance measures for the compared models for dataset 2.

| **Proposed Models** | | | | | |
|---|---|---|---|---|---|
| **Criteria** | **WSB** | **GA–LSTM** | **GA–RF** | **GA–SVR** | **GA–XGBoost** |
| MAE (kWh) | 464.93 | 466.92 | 529 | 752.50 | 567.83 |
| sMAPE (%) | 18.72 | 18.80 | 21.57 | 32.68 | 23.36 |
| Traditional Time Series Models | | | | | |
| **Criteria** | **Exponential Smoothing** | **Double Exponential Smoothing** | **Holt-Winters Additive** | **Holt-Winters Multiplicative** | **ARIMA** |
| MAE (kWh) | 771.36 | 1158.58 | 867.29 | 668.62 | 774.48 |
| sMAPE (%) | 30.42 | 41.59 | 33.26 | 26.77 | 30.47 |
| Other Artificial Neural Network Models | | | | | |
| **Criteria** | **GA–CNN** | **GA–ESN** | | | |
| MAE (kWh) | 633.16 | 681.58 | | | |
| sMAPE (%) | 25.68 | 28.62 | | | |

In contrast to the 5-fold cross-validation results, the GA–LSTM model demonstrated superior performance in electricity consumption forecasting for dataset 1 data in a 12-month-ahead horizon. With an sMAPE of 15.35% and an MAE of 2264.50, the GA–LSTM model significantly surpassed the GA–RF model, which achieved an sMAPE of 21.83% and an MAE of 3211.33. Moreover, the GA–LSTM model also outperformed traditional time-series forecasting models, with exponential smoothing ranking as the second-best performer, yielding an MAE of 3207.78 and an sMAPE of 21.82%.

Applying the WSB approach with a GA–LSTM model and the weak predictors GA–RF, GA–SVR, and GA–XGBoost resulted in a significant improvement in overall model performance, with an sMAPE of 13.90% and MAE of 1990.87. This demonstrates that the cooperative ensemble learning model is more effective for this specific problem, outperforming individual models in terms of sMAPE and MAE.

The remaining proposed models exhibited relatively similar performance, with sMAPE values ranging from 21.83% to 24.45% and MAE values ranging from 3211.33 to 3760.75. In contrast, the performance of traditional time-series forecasting models and other artificial neural network models was, in most cases, significantly lower, with sMAPE values ranging from 21.82% to 43.31% and MAE values ranging from 3207.78 to 5415.74.

A similar pattern was observed in dataset 2, where the GA–LSTM model once again demonstrated the best performance in the testing phase, with an MAE of 466.92 and an sMAPE of 18.80%.By applying the WSB approach with the GA–LSTM, the overall model performance improved significantly, achieving an sMAPE of 18.72% and an MAE of 464.93. The GA–RF model also maintained its second-best position, achieving an MAE of 529.00 and an sMAPE of 21.57%. Both models outperformed the performance of traditional time-series forecasting models and other artificial neural network models. The Holt-Winters multiplicative method, a traditional model, achieved the highest performance with an sMAPE of 26.77% and an MAE of 668.62. Among the other neural networks, the GA–CNN model demonstrated the best performance, with an sMAPE of 25.68% and an MAE of 633.16. Of the remaining models, the proposed GA–SVR and double exponential smoothing demonstrated the lowest level of performance, achieving sMAPE values of 32.68% and 41.59%, and MAE values of 752.50 and 771.36, respectively.

The discrepancy between the 5-fold cross-validation results and the 12-step-ahead forecast might be attributed to the occurrence of novel events within the time series, to which the models were not adequately exposed during training. Alternatively, the accumulation of forecast errors over successive predictions could have contributed to the observed divergence.

Figure 10 provides a visual comparison of estimated energy consumption for the IFPR–Palmas Campus, as predicted by each model, against the actual observed values throughout the year 2023. This graphical representation effectively underscores the predictive patterns exhibited by the different models.

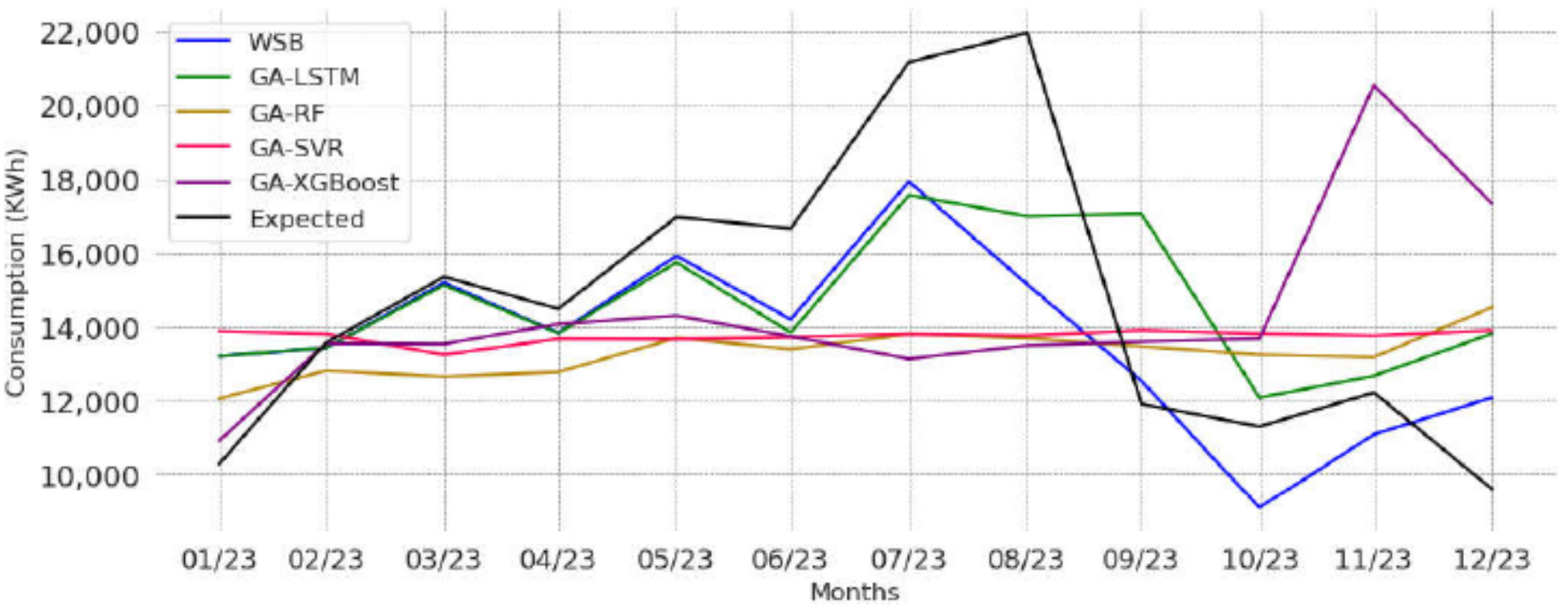

**Figure 10.** Comparison of performance between WSB, GA–LSTM, GA–RF, GA–SVR, and GA–XGBoost for electricity consumption in 12-months-ahead (Jan 2023–Dec 2023) at Palmas Campus (dataset 1).

A visual analysis indicates that all models demonstrated inaccuracies in predicting the sustained peak in energy consumption observed between June and September 2023. Among the models, the GA–LSTM model exhibited the closest behavior to the actual trend during this period. This anomalous data pattern suggests the occurrence of an unforeseen event, not present in the training dataset, that significantly influenced electricity consumption during this period.

The GA–SVR model exhibited a high degree of stability, consistently underestimating both the amplitude and frequency of data fluctuations and trends. Conversely, the GA–XGBoost model began forecasting in a highly efficient manner, outperforming others in the initial months. However, as errors accumulated, it demonstrated excessive sensitivity to data variations, resulting in highly unstable predictions characterized by frequent reversals in direction.

The GA–RF and GA–XGBoost models initially demonstrated proficiency in capturing upward trends within the time series. However, from March 2023 onwards, the models' performance declined as they exhibited increased resistance to data fluctuations, converging towards a more stable pattern resembling the behavior of the GA–SVR model.

The GA–LSTM model demonstrated a strong ability to track both upward and downward trends within the time series, consistently maintaining its performance throughout the entire period. However, similar to the other evaluated models, the models had difficulties capturing the energy consumption peak observed between June and September 2023.

As depicted in Figure 11, the electricity consumption forecasting results for IFPR–Coronel Vivida Campus also proved challenging to model. A peak in electricity consumption was observed between January 2023 and July 2023, similar to the IFPR–Palmas Campus time series. All models exhibited inaccurate behavior during this period.

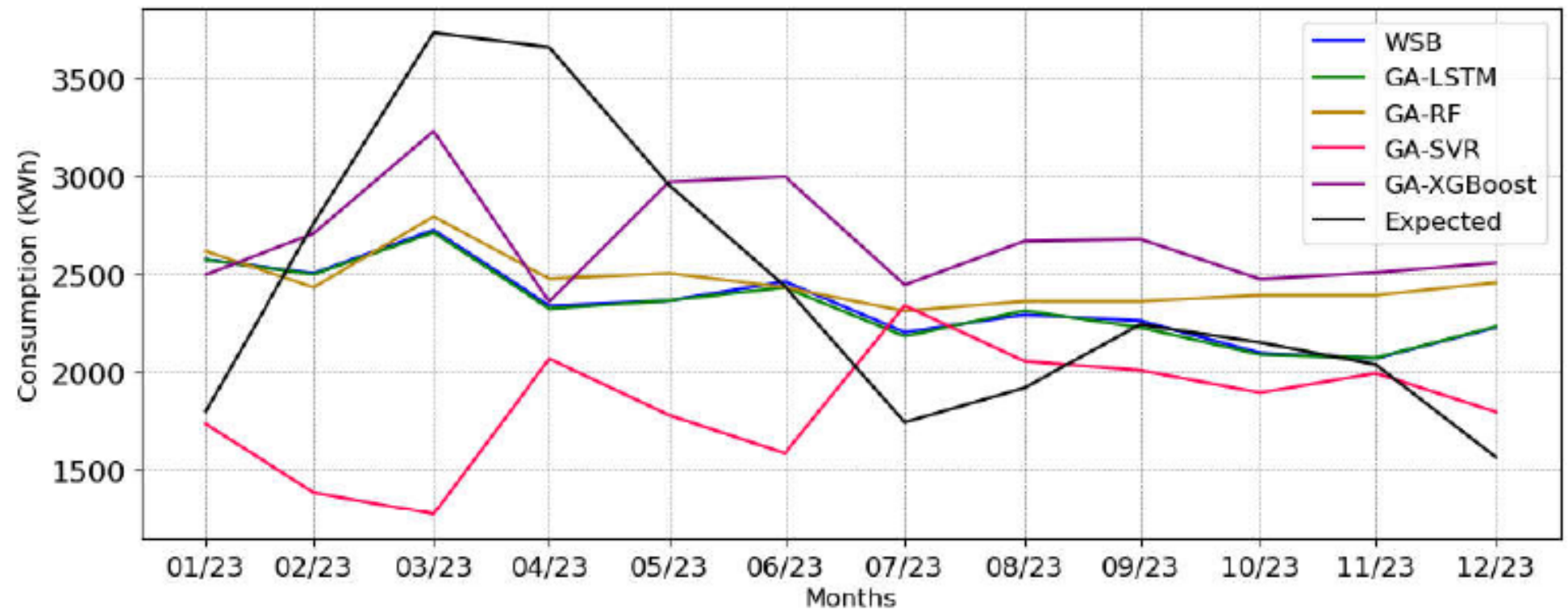

**Figure 11.** Comparison of performance between WSB, GA–LSTM, GA–RF, GA–SVR and GA–XGBoost for electricity consumption in 12-months-ahead (Jan 2023–Dec 2023) at Coronel Campus (dataset 2).

In this instance, the GA–SVR model exhibited inconsistent and unstable forecasting behavior. Despite this new behavior, it failed to adequately fit the data and performed the worst among the proposed models. Similarly, the GA–XGBoost model also experienced a decline in performance.

The GA–LSTM and GA–RF models exhibited the closest alignment with real-world data, with the GA–LSTM being the most suitable for the dataset. However, even these models struggled to accurately capture the peak consumption period between January and July 2023. This limitation underscores the challenge of identifying and responding to anomalous events within time-series data.

Additionally, we explored advanced neural networks in the field of forecasting, such as ESN and CNN, in an attempt to achieve error rates below 10%. Unfortunately, our attempts were unsuccessful, reinforcing the complexity of electricity consumption modeling, as previously noted in the literature.

In our proposed approach, the goal is to obtain 12-month-ahead forecasts using a recursive strategy. This method involves using forecasts from previous months as inputs for predicting the next month. As a result, any errors associated with each forecast accumulate throughout the process, increasing the percentage error by the end. Moreover, to simulate real-world conditions, the model stores its predictions and updates subsequent values based on these forecasted values. Consequently, if the model generates significant errors at any point, subsequent forecasts will be directly impacted. This is one of the reasons why the errors exceed 10%.

Furthermore, the complexity of electricity consumption modeling has been emphasized in the literature. For example, in [13], models with a 3-month forecast horizon showed MAPE errors as high as 31%. Similarly, for one-step-ahead forecasting in [15], the performance of the compared models ranged between 5.9% and 11.3%.

*4.2. SHAP Analysis*

Interpretability is essential for fostering trust and transparency in artificial intelligence systems. SHAP values provide invaluable insights into model behavior by quantifying the contribution of each feature to a specific prediction. Beyond feature selection, SHAP values elucidate the magnitude and direction of feature influence, enhancing model comprehension and explainability.

To identify the key determinants of energy consumption at IFPR and to optimize resource allocation, a SHAP analysis was conducted on the GA–RF model, given its best performance among those suitable for SHAP analysis in this study. This approach enabled the quantification of each variable contribution to the model predictions [21].

SHAP values are commonly visualized graphically to provide insights into the magnitude and direction of each feature's impact on the model's predictions. The arrows within each plot represent the impact of individual features on the final prediction. Positive values (rightward red arrows) indicate features that increase the predicted value, whereas negative values (leftward blue arrows) decrease it. The cumulative effect of these feature contributions determines the overall prediction.

Regarding the Palmas Campus, the GA–RF model predicted an average electricity consumption of 15,441.89 kWh. Each feature included in the model contributed to this prediction, either increasing or decreasing the forecast value. Figure 12 presents the average values for each feature and their corresponding contributions.

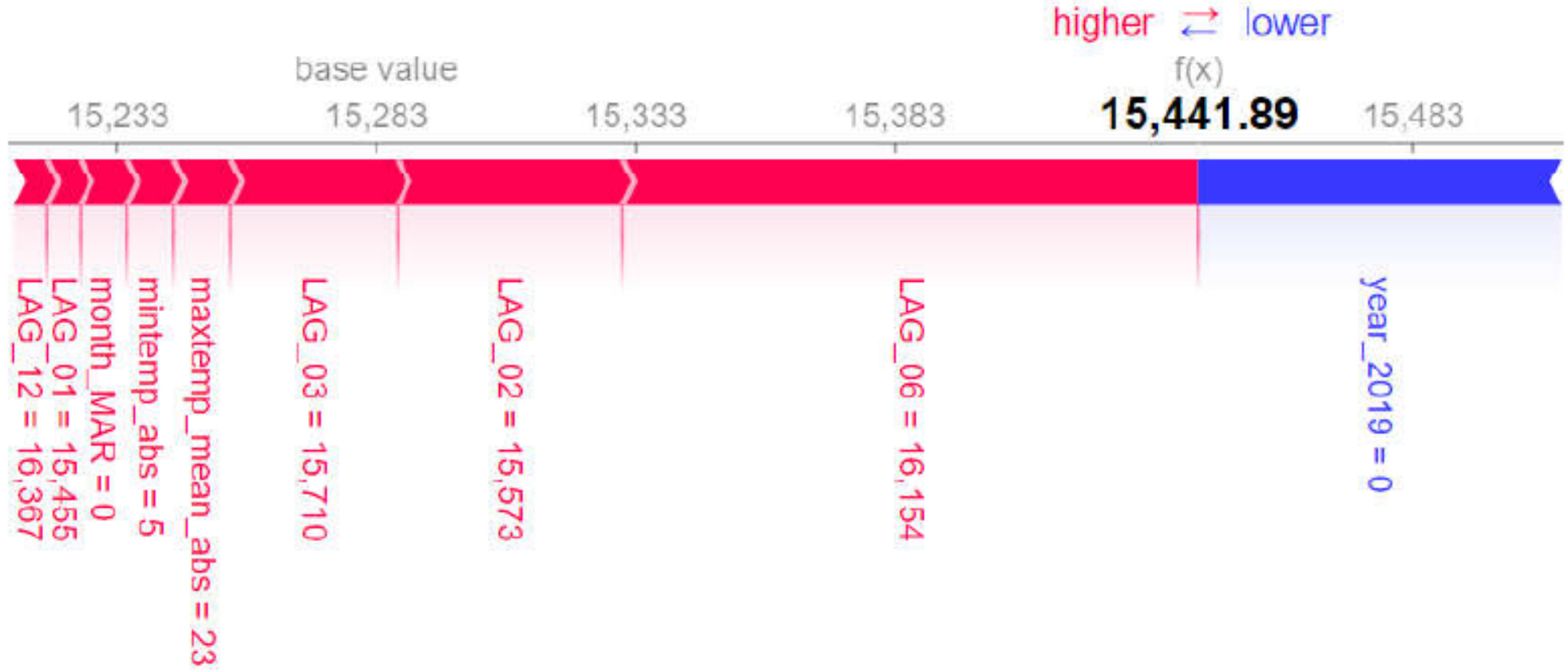

**Figure 12.** Force plot of average SHAP values from GA–RF trained on Palmas Campus data. The values next to each feature represent its mean value across all evaluated predictions in SHAP.

The prominent SHAP values associated with Lag-06, along with the substantial contributions of Lag-02 and Lag-03, underscore a strong dependence of electricity consumption on its historical values. These findings align with the results of the KPSS test, which confirmed the presence of autocorrelation in the initial lags of the time series.

The significant impact of the "year 2019" variable suggests a notable deviation in electricity consumption patterns for that year. The variable's importance to the model performance indicates that its exclusion would have negatively impacted the accuracy of predictions. This finding highlights the need for further investigation into the factors contributing to this phenomenon.

Among the climate variables, "absolute average maximum temperature" and "absolute minimum temperature" exhibited a positive correlation with energy consumption. Although these variables were of lesser importance compared to others, the findings suggest a relationship between temperature variations and electricity consumption in the IFPR–Palmas Campus region, especially considering its colder climate. However, further research is needed to confirm this hypothesis.

In the Coronel Vivida Campus, the GA–RF model predicted an average electricity consumption of 2164 kWh. The distribution of feature importance and direction differed significantly from the results obtained for the Palmas Campus, as illustrated in Figure 13.

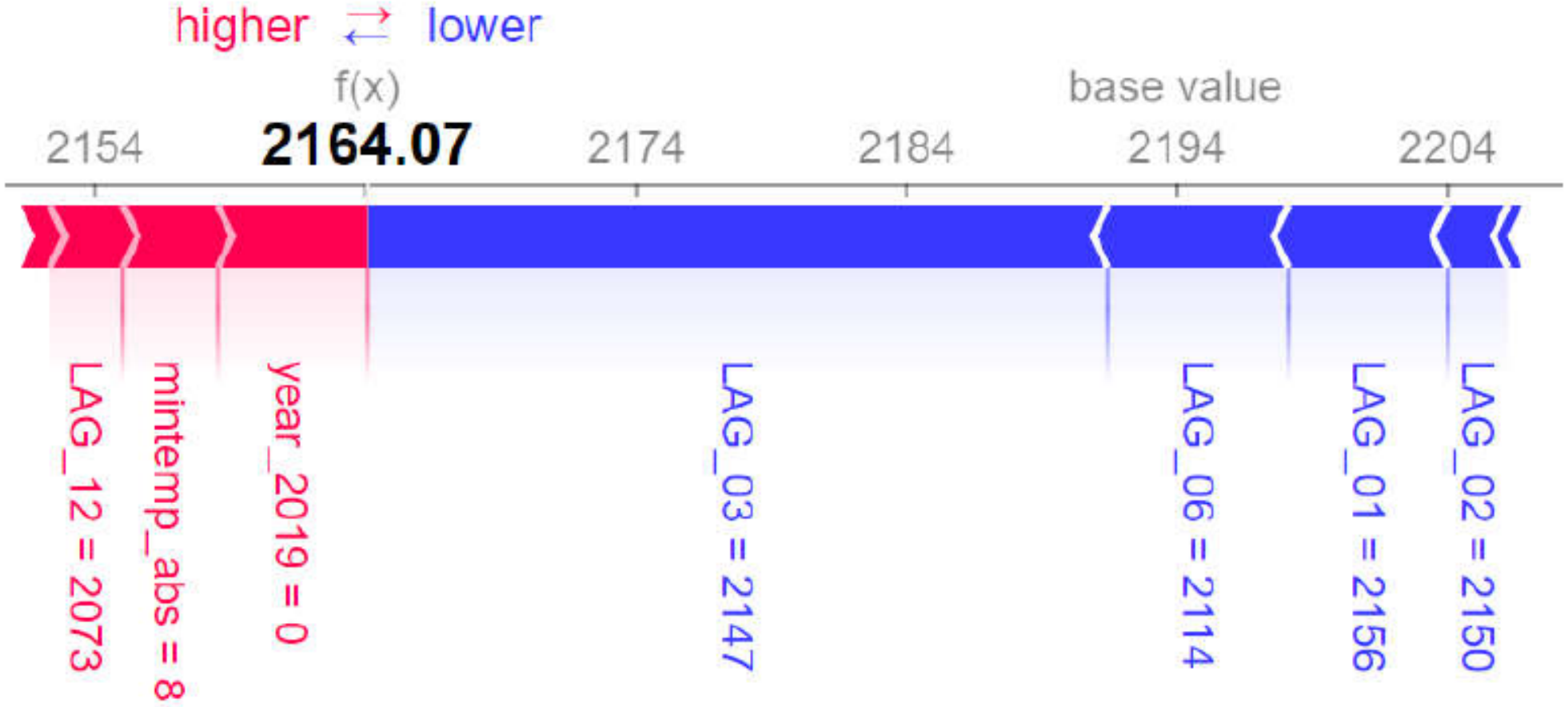

**Figure 13.** Force plot of average SHAP values from GA–RF trained on Coronel Vivida Campus data. The values next to each feature represent its mean value across all evaluated predictions in SHAP.

In this analysis, the SHAP values for features "Lag-03", "Lag-06", and "Lag-01" exhibited the highest contributions to the electricity consumption forecast. Unlike the Palmas Campus, where these lagged values had a positive impact, in Coronel Vivida, they negatively influenced the forecast. This implies that higher lagged consumption values tend to decrease the predicted consumption for the current period, suggesting an inverse relationship between past and present consumption at this campus.

The "year-2019" feature once again proved to be a significant predictor in the model, this time exerting a positive influence. This suggests that this feature is increasing the predicted value of electricity consumption.

The climatic feature "absolute minimum temperature" had a minimal contribution to the model. Although it was the most influential among the climatic variables incorporated in the Coronel Vivida Campus, its overall impact was relatively small. Similar to the Palmas Campus, this variable also contributed positively to the forecast. This suggests a comparable influence of regional climate on electricity consumption and warrants further investigation.

## 5. Conclusions and Future Work

The inherent variability and complex patterns within educational institutions electricity consumption data presented a significant challenge for predictive modeling. However, the proposed approach, employing cooperative ensemble learning with hyperparameter optimization, SHAP-based feature selection, and a Weaker Separator Booster, demonstrated promising potential in addressing this complexity.

The elimination of less influential features through feature selection significantly reduced model complexity and enhanced performance. By leveraging SHAP values derived from XGBoost and RF models, we quantified the contribution of each feature to the prediction and subsequently removed those with minimal impact. This analysis led to the removal of 36 variables from the original dataset of 47 predictors.

GA and PSO were instrumental in identifying optimal hyperparameter configurations for the LSTM, RF, SVR, and XGBoost models. By leveraging GPU-accelerated parallel processing, the computational efficiency of evaluating numerous hyperparameter combi-nations was significantly enhanced. Notably, the GA consistently outperformed PSO in achieving optimal function values within a shorter timeframe.

A comparative analysis of GA–LSTM, GA–RF, GA–SVR, and GA–XGBoost models for 12-month-ahead electricity consumption forecasting at a university revealed the superior performance of the GA–LSTM model for both the IFPR–Palmas and IFPR–Coronel Vivida campuses. The GA-LSTM model achieved the lowest sMAPE of 15.35% and MAE of 2264.50 for the Palmas dataset and an sMAPE of 18.80% and MAE of 771.36 for the Coronel Vivida dataset. Compared to traditional time-series forecasting models like exponential smoothing, Holt-Winters, ARIMA, and other neural network architectures such as ESN and CNN, the GA–LSTM model also exhibited superior performance.

Utilizing the Weaker Separator Boosting (WSB) approach with the GA–LSTM model as the strongest learner, and GA–RF, GA–SVR, and GA–XGBoost as weaker learners, led to a notable enhancement in model performance across both the Campus Palmas (sMAPE: 13.90%, MAE: 1990.87) and Campus Coronel Vivida (sMAPE: 18.72%, MAE: 464.93) datasets. It is important to acknowledge that the model performance can be substantially affected by the selection of hyperparameters and ensemble weights. Future research should investigate alternative strategies for combining the weaker predictors, as well as explore variations in parameters such as the boosting function and weight settings, to assess the broader applicability of the WSB approach in different domains.

The analysis of SHAP values provided insights into the most important features for predicting electricity consumption in each academic campus. Furthermore, understanding the direction of this influence, whether positive or negative, offered valuable insights into the factors that most significantly affect electricity consumption. In the time series analyzed, lagged values of the series themselves were the most relevant for prediction, exhibiting both positive and negative influences. In contrast, climatic variables had a significantly smaller impact on the forecast, as evidenced by their lower SHAP values. Some temporal dummy variables (month and year) also showed significant contributions, indicating the need for further investigation into these periods to identify potential events that influenced energy consumption.

In future work, we suggest expanding this methodology to include electricity consumption data from various universities located in geographically diverse regions. We aim to explore the incorporation of novel exogenous variables specific to educational institutions, with the objective of evaluating the individual contributions of these variables to forecast accuracy. Furthermore, considering the cumulative forecast errors observed in all models when using a 12-month forecast horizon, which may be influenced by the interdependence of forecasts, we intend to assess the performance of the models with shorter forecast horizons.

## Abbreviations

The following abbreviations are used in this manuscript:

| | |
|---|---|
| ACF | Autocorrelation Function |
| ANN | Artificial Neural Network |
| ARIMA | Autoregressive Integrated Moving Average |
| CNN | Convolutional Neural Network |
| CV | Coefficient of Variation |
| DTO | Dipper Throated Optimization |
| ELM | Extreme Learning Machines |
| ESN | Echo State Network |
| EWO | Earth Worm Optimization |
| GA | Genetic Algorithm |
| GPU | Graphics Processing Unit |
| GRU | Gated Recurrent Units |
| GWO | Grey Wolf Optimizer |
| IDR | Paraná Institute of Rural Development |
| IFPR | Federal Institute of Paraná |
| KNN | K-Nearest Neighbors |
| KPSS | Kwiatkowski–Phillips–Schmidt–Shin |
| LMSR | Linear Model Stepwise Regression |
| LR | Linear Regression |
| LS | Least Squares |
| LSSVR | Least Squares Support Vector Regression |
| LSTM | Long Short-Term Memory |
| kWh | Kilowatt Hour |
| MAE | Mean Absolute Error |
| MAPE | Mean Absolute Percentage Error |
| MSE | Mean Squared Error |
| MetaFA | Metaheuristic Firefly Algorithm |
| ML | Machine Learning |
| MLP | Multilayer Perceptron |
| NB | Naive Bayes |
| PACF | Partial Autocorrelation Function |
| PSO | Particle Swarm Optimization |
| R | Correlation Coefficient |
| RF | Random Forest |
| RMSE | Root Mean Squared Error |
| RT | Regression Tree |
| SARIMA | Seasonal Autoregressive Integrated Moving Average |
| SD | Standard Deviation |
| SHAP | Shapley Additive Explanations |
| SIMEPAR | Paraná Meteorological System |
| sMAPE | Symmetric Mean Absolute Percentage Error |
| SVM | Support Vector Machine |
| SVR | Support Vector Regression |
| XGBoost | Extreme Gradient Boosting |
| WSB | Weaker Separator Booster |